\documentclass[journal, twoside, web]{ieeecolor}
\usepackage{generic}
\usepackage{amsmath,amssymb,amsfonts}
\usepackage{graphicx}
\usepackage{subfig}
\usepackage{adjustbox}
\usepackage{pgf}
\usepackage{soul}
\usepackage{algorithm}
\usepackage[noend]{algpseudocode}
\usepackage{booktabs}
\usepackage{tcolorbox}
\usepackage{xcolor}
\usepackage{multicol}
\usepackage{multirow}
\usepackage[hidelinks]{hyperref}
\usepackage{float}
\usepackage{rotating}
\usepackage{pgfplots}
\usepackage{lscape}
\usepackage{tikz}
\usetikzlibrary{positioning}

\def\BibTeX{{\rm B\kern-.05em{\sc i\kern-.025em b}\kern-.08em
    T\kern-.1667em\lower.7ex\hbox{E}\kern-.125emX}}
\begin{document}
\title{A Cross Attention Approach to Diagnostic Explainability Using Clinical Practice Guidelines for Depression}

\author{Sumit Dalal\textsuperscript{*}, Deepa Tilwani\textsuperscript{*}, Manas Gaur, Sarika Jain, Valerie L. Shalin, and Amit P. Sheth
\thanks{*Sumit Dalal and Deepa Tilwani contributed equally to this work as first authors.}
\thanks{Sumit Dalal is with the National Institute of Technology Kurukshetra, Haryana, India 136119 (e-mail: sumitdalal9050@gmail.com).}
\thanks{Deepa Tilwani is with the Artificial Intelligence Institute at the University of South Carolina (AIISC), Columbia, USA (e-mail: dtilwani@mailbox.sc.edu).}
\thanks{Manas Gaur is with the University of Maryland, Baltimore County, USA (e-mail: manas@umbc.edu).}
\thanks{Sarika Jain is with the National Institute of Technology Kurukshetra, Haryana, India 136119 (e-mail: jasarika@nitkkr.ac.in).}
\thanks{Valerie Shalin is with Wright State University, USA (e-mail: valerie.shalin@wright.edu).}
\thanks{Amit P. Sheth is with the Artificial Intelligence Institute at the University of South Carolina (AIISC), Columbia, USA (e-mail: amit@sc.edu).}}

\maketitle

\begin{abstract}
The lack of explainability in using relevant clinical knowledge hinders the adoption of artificial intelligence-powered analysis of unstructured clinical dialogue. A wealth of relevant, untapped Mental Health (MH) data is available in online communities, providing the opportunity to address the explainability problem with substantial potential impact as a screening tool for both online and offline applications. Inspired by how clinicians rely on their expertise when interacting with patients, we leverage relevant clinical knowledge to classify and explain depression-related data, reducing manual review time and engendering trust. We developed a method to enhance attention in contemporary transformer models and generate explanations for classifications that are understandable by mental health practitioners (MHPs) by incorporating external clinical knowledge. We propose a domain-general architecture called \textbf{P}roces\textbf{S} knowledge-infused cross \textbf{AT}tention ({\fontfamily{cmss}\selectfont PSAT}) that incorporates clinical practice guidelines (CPG) when computing attention. We transform a CPG resource focused on depression, such as the Patient Health Questionnaire (e.g. PHQ-9) and related questions,  into a machine-readable ontology using SNOMED-CT. With this resource,  {\fontfamily{cmss}\selectfont PSAT} enhances the ability of models like GPT-3.5 to generate application-relevant explanations. Evaluation of four expert-curated datasets related to depression demonstrates  {\fontfamily{cmss}\selectfont PSAT}'s application-relevant explanations. {\fontfamily{cmss}\selectfont PSAT} surpasses the performance of twelve baseline models and can provide explanations where other baselines fall short. 

\end{abstract}


\section{Introduction}
\label{sec:introduction}

\IEEEPARstart{A}{s} the number of mental illness cases in the US rises, expenditure on AI for mental health (MH) services has grown from \$200M in 2020 to \$375M in 2022 \cite{samhsa2023research}. Nevertheless, the adoption of AI  tools for MH services remains modest ($<$20\%) \cite{nadeem2023pewresearch}. Nearly all mental health practitioners (MHPs) are skeptical about the role of AI in the automated screening of patients with MH conditions \cite{cushing2023health}. 

According to  Kelly et al.'s highly cited article \cite{kelly2019key}, clinicians must understand how proposed AI systems align with current clinical practice guidelines (CPGs). However, current AI methods either do not follow these guidelines or fail to explain their compliance. Although Language Models (LMs) hold great potential as assistive tools for MHPs, they are opaque BlackBoxes. Efforts to explain LM decisions such as LIME \cite{lundberg2017unified} and SHAP \cite{ribeiro2016should} rely on post-processing steps for an explanation.

\begin{table*}[!ht]
    \resizebox{\textwidth}{!}{
    \begin{tabular}{p{8cm}|p{8cm}}
    \toprule[2pt]
        MentalBERT & {\fontfamily{cmss}\selectfont PSAT} (Proposed Approach) \\ \toprule[2pt]
        Lately I've been [...]\underline{depression} but I've never been to \underline{therapy} because I couldn't afford it [...] Now I live on my own in another city. Yesterday I discovered that my university provides \underline{psychological} \underline{help} for students for free. Do you think I should give it a go? […] I know they don't provide help for very \underline{serious issues} (you'll need a \underline{psychiatrist} for that) and I hope they don't take care of only ``university related \underline{problems}".On the other hand, I have nothing to \underline{lose} because it's free. Did you ever tried anything like that? & 
        Lately I've been \textbf{feeling really low}. I \textbf{can't make myself leave the bed}, I  \textbf{start crying out of the blue} and everything is just  \textbf{so heavy} [...]  \textbf{depression} but I've  \textbf{never been to therapy} because I couldn't afford it on [...] Now I live on my own in another city. Yesterday I discovered that my university provides  \textbf{psychological help} for students for free. Do you think I should give it a go? [...] I know they don't provide help for very  \textbf{serious issues} (you'll need a psychiatrist for that) and I hope they don't take care of only university related ``problems".  [...] \\ \midrule[1pt]
        PHQ-9 Question Answered: Q2 & PHQ-9 Question Answered: Q1, Q2, Q3, Q4, Q9 \\ \midrule[1pt]
        Predicted Probability ($f(\hat{y}|x, \theta)$): 0.54 & Predicted Probability ($f(\hat{y}|x, \theta_{\mbox{PHQ-9}})$): 0.72 
        \\ \bottomrule[1pt]
    \end{tabular}}
    \caption{A comparison between MentalBERT (an existing LM) and {\fontfamily{cmss}\selectfont PSAT} on the grounds of application-relevant explainability as seen from the lens of PHQ-9. The objective is to enhance the LM's predictive accuracy and ability to provide MHP-understandable explanations. Unlike the underlined syntactic phrases detected by MentalBERT, the concepts highlighted by {\fontfamily{cmss}\selectfont PSAT} are centered around depression and the PHQ-9, as shown in Figure 3.}
    \label{tab:1}
\end{table*}

We develop an approach to achieve application-relevant explainability that leverages CPGs used by MHPs. Our motivation is governed by the research question: \textit{Is it possible to design the LM architectures to be inherently explainable in terms of CPGs and their constituents’ clinical terms and concepts?}

We propose ProcesS knowledge-infused cross ATtention ({\fontfamily{cmss}\selectfont PSAT})\footnote{This paper has been accepted for publication in IEEE Journal of Biomedical and Health Informatics}, a domain-general LM architecture that incorporates CPGs through a novel, inherently explainable cross-attention mechanism. {\fontfamily{cmss}\selectfont PSAT}'s cross-attention ensures that the LM pays attention to clinically relevant concepts and highly important words or phrases when making decisions. CPG is, therefore, causal in the reasoning process.

\begin{figure}[t]
\includegraphics[width=0.46\textwidth]{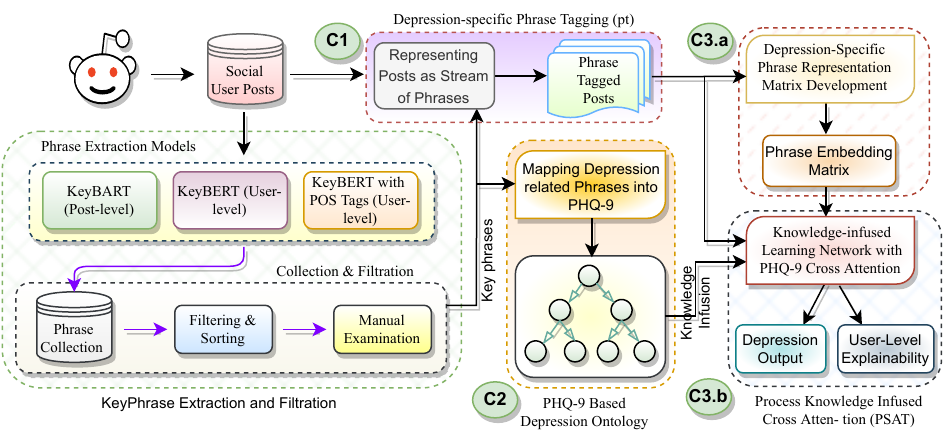}
\caption{Phrase Extraction and Other Resource Generation for Explainable Depression Detection. The red circles containing C1, C2, and C3 denote our contributions. C1 represents the adaptation of the CLEF eRisk dataset; C2 shows the development of PHQ-9-based depression ontology, C3.a displays part 'a' our third contribution of this work, which is the development of depression-specific phrase embedding matrix, and C3.b is the knowledge-infused cross-attention network for explainable depression detection.}
\label{fig:workflow}
\end{figure}

We focus on depression and use the Patient Health Questionnaire (PHQ-9) as the CPG. Using three gold-standard datasets for demonstrating explainability in MH, we provide three main contributions: First, we introduce and illustrate {\fontfamily{cmss}\selectfont PSAT}, describing the architectural underpinnings, designed to achieve explainability by \textit{enhancing the attention} of transformer models with access to CPG. Second, we provide a novel metric called Average Knowledge Capture (AKC) that guides a quantitative evaluation of {\fontfamily{cmss}\selectfont PSAT} outputs relative to CPGs. Third, we conduct a qualitative evaluation of {\fontfamily{cmss}\selectfont PSAT}, 
showing the benefit of its attention mechanism in improving the quality of explanations compared to the conventional approach of generating explanations through prompting in GPT 3.5. Our products from this research are the {\fontfamily{cmss}\selectfont PSAT} and the computationally accessible PHQ-9 ontology. The workflow of the proposed {\fontfamily{cmss}\selectfont PSAT} model, in Figure \ref{fig:workflow} highlights the aforementioned contributions in this research.

\textit{Generalizability:} From a methodological standpoint, although this approach is informed by a CPG resource for depression, it remains broadly applicable, needing only a computationally accessible representation of the relevant knowledge. By decoupling the content from the model's architecture, we can easily adapt this method to other domains, such as Anxiety, using resources like the Generalized Anxiety Disorder Questionnaire (GAD-7) \cite{GAD} and Suicide Ideation-Behavior-Attempt using Columbia Suicide Risk Severity Rating Scale (CSSRS) \cite{posner2011columbia}. Table \ref{tab:1} demonstrates two advantages of the proposed {\fontfamily{cmss}\selectfont PSAT} over MentalBERT: {\fontfamily{cmss}\selectfont PSAT} identifies all salient words or phrases that can answer questions in the PHQ-9, and as it learns to focus on these pertinent elements, it improves prediction confidence. In the following sections, we describe our proposed solution, the {\fontfamily{cmss}\selectfont PSAT} method, and provide an in-depth analysis showcasing {\fontfamily{cmss}\selectfont PSAT}'s ability to achieve application-relevant explainability.

\textbf{NOTE:} Examples in this manuscript are potentially disturbing and have been subject to ellipsis and paraphrasing to reduce exposure to negative content. Additional examples, links to the dataset, and reproducibility code are available \href{https://github.com/Deepa-Tilwani/DepressionDetection}{\fontfamily{cmss}\selectfont PSAT-GitHub}.

\section{Related Work}
\noindent \textit{Linking to Application-Relevant Explainability:} Optimizing explanations from LMs for human usage remains challenging \cite{samek2019towards}. Previous efforts to explain BlackBox LMs have employed post-hoc methods such as surrogate models (e.g., LIME and SHAP \cite{scodari2023using}), visualization techniques \cite{vig2019bertviz,gunaratna2022explainable}, and local perturbations \cite{la2021guaranteed}. These provide meta-explanations but only at a low level of abstraction. \textit{Post-hoc Explainability} represents a simple approach to explaining LMs for users. This method entails interpreting the attention mechanism of the LM and associating it with concepts present in an ancillary knowledge source. This approach relies on two methods: (a) network dissection \cite{netdissect2017} and (b) entity linking-based approaches using knowledge bases \cite{chen2023towards}. The latter method benefits from vast healthcare knowledge bases like UMLS (Unified Medical Language System), SNOMED-CT (Systematized Nomenclature of Medicine-Clinical Terms), and ICD-10 (International Classification of Diseases) \cite{karim2023biomedical,gaur2019knowledge}. While post-hoc methods offer explainability, they are not participants in the reasoning process and do not constrain the underlying mechanisms to prioritize essential elements consistently. The explanation could simply be epiphenomenal. 

Recent focus has shifted to explainability by design, particularly for critical domains such as healthcare, \cite{holzinger2019causability}, biology \cite{huang2023global}, and others \cite{gade2020explainable}. Joyce et al.'s hypothetical framework of Transparency and Interpretability For Understandability (TIFU) readily connects with the notion of explainability by design for MHPs \cite{joyce2023explainable}. The proposed {\fontfamily{cmss}\selectfont PSAT} model is inspired by the TIFU framework and significantly contributes to knowledge infusion using CPGs.


\noindent \textit{Mental Health:}  Concentrated effort over the last decade endeavors to enhance MH services by applying AI and machine learning to both social media and clinical data \cite{zhang2022natural,sigman2021artificial}. These efforts encompass a variety of tasks, such as identifying and categorizing MH issues \cite{liu2023improving}, assessing the severity of MH conditions \cite{gaur2019knowledge,naseem2022early}, answering questions related to MH \cite{gupta}, and engaging in conversational interactions \cite{sarkar2023towards}. While LMs perform well on these tasks, their BlackBox nature compromises trust, 
due to the absence of MHP-desired explainability \cite{semanticsofblack}. Thus, MH LMs are attempting to incorporate clinical instruments like questionnaires and role-playing based on actual clinical diagnostic interviews \cite{demasi-etal-2019-towards} or involve MHPs in the preparation of annotation guidelines \cite{garg2022cams,shing2018expert}. These resources serve as the foundation for enhancing the explainability and practicality of MH models. Utilizing neural architectures that accept domain-specific knowledge bases (KBs) and are therefore adaptive, incorporating the underlying knowledge from these resources and data, has significant potential to bolster MH-specific downstream tasks.

\noindent \textit{Knowledge Infusion:} 
Current knowledge infusion methods fall into four groups: (A) combining knowledge embeddings with input embeddings during model training (e.g., ERNIEv1 \cite{zhang2019ernie}, ERNIEv2 \cite{sun2020ernie}, KALA \cite{kang2022kala}), (B) enhancing input embeddings with knowledge using dense passage retrieval (e.g., RAG \cite{lewis2020retrieval}, ISEEQ \cite{gaur2022iseeq}), (C) customizing and optimizing the attention function of LMs using both input and knowledge semantics (KI-BERT \cite{faldu2021ki}, ELMO \cite{peters2019knowledge}), and (D) using datasets grounded in additional lexicons or knowledge-graphs. Recent studies have utilized CPGs to examine the fidelity of BlackBox LMs' attention on tasks like depression classification and suicide risk severity detection \cite{nguyen2022improving, mahbub2023cpgqa, zirikly2022explaining, roy2022process2}. These studies belong to category (D), which suffers from a reliance on manual annotation. It is time-consuming to achieve gold-standard inter-annotator agreements, costly, and error-prone.  We advance the research along the category (C)  by infusing CPGs within an LM's computing machinery to achieve application-relevant explainability by design.

\section{\fontfamily{cmss}\selectfont PSAT}

\begin{figure}[!ht]
\centering
\includegraphics[width=0.5\textwidth]{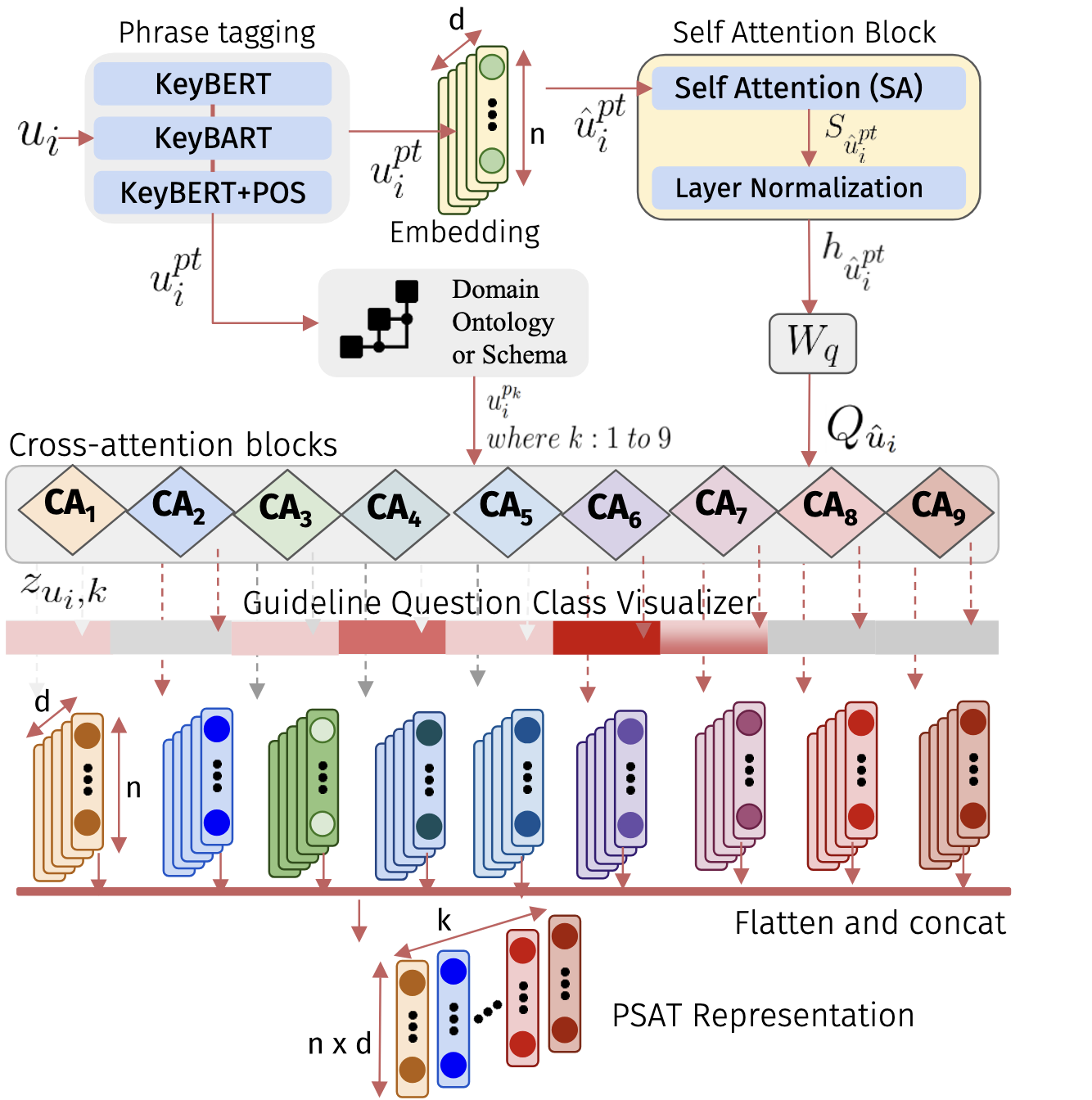}
\caption{Overview of {\fontfamily{cmss}\selectfont PSAT} model. Nine cross-attention (CA) blocks represent the Nine PHQ-9 questions. $n$ represents the number of topical phrases in a user document to map to the PHQ-9 ontology. $d$ is the embedding. {\fontfamily{cmss}\selectfont PSAT} allows visualization of PHQ-9-level attentions (represented in different colors) as application-relevant explanations useful for MHPs.}
\label{fig:cross-attn-network}
\end{figure}

\subsection{Architecture} {\fontfamily{cmss}\selectfont PSAT} builds on the contemporary BERT architecture, known for its superior simplicity and effectiveness. This also allows for meaningful comparisons with MentalBERT and other BERT-based MH-specific models. However,  {\fontfamily{cmss}\selectfont PSAT} is unique by including new cross-attention blocks. These blocks are different from traditional cross-attention blocks in decoder-only LMs and enable the model to compute attention between words from the text, phrases from the KB (such as PHQ-9 Depression ontology), and the questions in CPGs. The number of cross-attention blocks corresponds to the number of questions in the CPGs. Figure \ref{fig:cross-attn-network} shows nine cross-attention blocks representing the nine questions in the PHQ-9 questionnaire, which focuses on depression. 

{\fontfamily{cmss}\selectfont PSAT} is also supported by two external modules: \textit{Knowledge Infusion using PHQ-9 Depression Ontology}\footnote{The current architecture of {\fontfamily{cmss}\selectfont PSAT} uses an ontology. Other forms of knowledge can also be explored.} and \textit{PHQ-9 Class Explanation Visualizer}. Figure \ref{fig:cross-attn-network} shows the architecture of {\fontfamily{cmss}\selectfont PSAT} with the PHQ-9 Depression Ontology. 

\noindent The \textbf{PHQ-9} \textbf{D}epression \textbf{O}ntology (PHQ-9-DO) is a structured knowledge source wherein every class represents a question in PHQ-9 (see figure (\ref{fig:explanations})). The ontology is constructed as follows: (a) We leverage an existing PHQ-9 lexicon in \cite{yazdavar2017semi}. Each category represents a question from PHQ-9, and concepts per category are extracted from UMLS, SNOMED-CT, MedDRA, and social media-based Urban Dictionary. For creating the PHQ-9-DO, we transformed the lexicon into a specific format: $<category, words, \text{SNOMED-CT ids}>$ and gathered \textit{ids} using SNOMED-CT APIs \cite{carlsen2023snomedct}. An instance in PHQ-9-DO appears as follows $<Q1, \text{loss of pleasure}, 404684003>$.

The words in PHQ-9-DO were expanded by applying neural keyphrase extraction techniques across all posts in the PRIMATE, CLEF e-Risk, and CAMS datasets. For this purpose, we employed KeyBERT \cite{grootendorst2020keybert}, KeyBART \cite{kulkarni2022learning}, and an extended version of KeyBERT that incorporates part-of-speech tagging and TF-IDF \cite{danesh2015sgrank}. Consider the post presented in Table \ref{tab:1}. The keyphrases extracted using the above techniques encompass phrases like \textit{suffered depression, feeling low, need therapy, feeling low, need psychiatrist, psychological help service, crying out of the blue, can't leave the bed}, etc. This process generated approximately 18,000 keyphrases, filtered down to 4700 based on their presence in the SNOMED-CT KB. These keyphrases were aligned with the PHQ-9-DO by computing cosine similarity between each keyphrase and the classes in PHQ-9-DO with a cut-off score of 0.80. We considered key phrases as similar to those in the PHQ-9 Depression ontology created using the Unified Medical Language System, SNOMED CT, MedDRA, and social media-based Urban Dictionary. We also included words in the Urban Dictionary to account for everyday language. We evaluated the similarity of everyday language to the PHQ-9 ontology by utilizing natural language processing techniques to compare the semantic similarity between phrases in the everyday language corpus and the clinical terminology. This process involved using cosine similarity measures and machine learning models trained on large language datasets to ensure accurate mapping and assessment.

\noindent \textbf{PHQ-9 Class Explanation Visualizer} visualizes the attention patterns in {\fontfamily{cmss}\selectfont PSAT} against each cross-attention block. The attention weights reflect the model's internal representation of the relationships. 
{\fontfamily{cmss}\selectfont PSAT} also represents dependencies between different words/phrases in the input and questions in the attention blocks. The visualizer identifies key parts of the input text regarding how the LM processed the information and performed  \textit{MH classification.} The end-to-end training in {\fontfamily{cmss}\selectfont PSAT} generates a contextualized representation that any classifier can use. The current version of {\fontfamily{cmss}\selectfont PSAT} employs \textit{feed-forward neural network} (\textit{FFNN}) for classification.

\subsection{Algorithm} 
~{\fontfamily{cmss}\selectfont PSAT} takes a textual paragraph or sentences represented as $u_i$ is original input text for the i-th sample, $u_i$   $\in \mathbb{R}^{m}$, where m is the number of tokens. It outputs a contextualized representation $z_{u_i}$ with respect to PHQ-9-DO. This representation is then utilized by a \textit{FFNN} classifier for the downstream task of detecting depression or causes of depression (see the  Datasets section). The process involves transforming $u_i$ into $u^{pt}_i$, where  $pt$ stands for phrase tagging. This tagged input is then matched or validated against the domain ontology to create $u_i^{pk}$, an intermediate representation of input data that has been both phrase-tagged and aligned with the domain ontology, ensuring that the embeddings used in the cross-attention mechanism are deeply contextualized with clinical knowledge from the PHQ-9. The processed input $u_i^{pk}$  is subsequently used to generate embeddings, which are crucial for the cross-attention mechanism in the model. Phrase tagging preserves the meaning of sentences following the default LM tokenizing process \cite{gu2021ucphrase}. The phrases were identified and tagged using phrases identified by KeyBERT, KeyBART, and KeyBERT + POS, as illustrated in Figure \ref{fig:cross-attn-network} above. $u_i^{pt} \in \mathbb{R}^{m}$, maintaining the same length   m but enriched with tags.  Embedding representation of the phrase-tagged input, $\hat{u}_i^{pt} \in \mathbb{R}^{m \times d}$, where d   is the embedding dimensionality (typically 768 for models like BERT).

Following that, $u^{pt}_i$ is subjected to a series of steps, including transforming it into an embedding and undergoing two standard operations: self-attention and normalization. These steps result in the creation of its latent representation\footnote{we use $h_{\hat{u}^{pt}_i}$ rather than $h_{u^{pt}_i}$ to signify hidden states}, denoted as $h_{\hat{u}^{pt}_i}$.

The computation of $Q$, $K$, and $V$ vectors in {\fontfamily{cmss}\selectfont PSAT} differs from the traditional self-attention method proposed by Vaswani et al\cite{vaswani2017attention}. Precisely, to calculate $Q$, we perform a linear projection of the input representation $h_{u^{pt}_i}$, which is derived from the single self-attention block before knowledge infusion (refer to Figure \ref{fig:cross-attn-network}). The query vectors from the self-attention block, denoted as $Q \in \mathbb{R}^{m \times d}$.
Subsequently, we compute nine $K$ and $V$ vectors using linear projections of the embeddings of each question in PHQ-9. These question embeddings are obtained through a separately defined embedding layer, denoted as $K^{phq}_k$ and $V^{phq}_k$, where $k = {1,2,...,9 }$.

We generated \textit{nine} hidden representations of the input by calculating the cross attention between the input vector and each of the \textit{nine} $Q$ and $V$ vectors, which correspond to the PHQ-9 questions ($z_{u_i}$), final hidden representation after cross-attention, $z_{ui} \in \mathbb{R}^{m \times 9}$. Subsequently, we concatenated these nine hidden representations. Depending on the task, we fed the resulting concatenated vector into a classification head to obtain the final classification label or vector, $\hat{y}_i(u_i)$. {\fontfamily{cmss}\selectfont PSAT} learns using the standard binary cross-entropy loss, which is represented as $L(\hat{y}_i(u_i), y_i(u_i))$, with a threshold $\varepsilon$ set to 0.75. 
Cross-attention in PSAT uses a domain-specific infusion of knowledge, which allows each attention head to access and integrate information from predefined clinical guidelines (like PHQ-9). It does not just focus on different input parts based on previously generated context (as in standard encoder-decoder models). Still, it is actively guided by an external, structured set of clinical concepts.  The purpose of using cross-attention in PSAT is to ensure that the analysis and predictions of the model are aligned with professional medical guidelines, thus ensuring that the model's outputs are both accurate and interpretable to health professionals.
By examining the individual cross-attention matrices, we can interpret which of the nine PHQ-9 questions contributed the most predictive value toward the final classification. Moreover, this illustrates that the attention mechanism in {\fontfamily{cmss}\selectfont PSAT} reflects the number of questions that can be addressed in a given post. Algorithm \ref{alg:PKiL} provides a formal pseudocode for the complete process within {\fontfamily{cmss}\selectfont PSAT}. Notably, line 7 and the associated clinical knowledge source, which at present is PHQ-9, can be changed based on the number of questions in CPGs used for assessing other MH conditions.

\begin{algorithm}[t]
\caption{\textbf{{\fontfamily{cmss}\selectfont PSAT}} - Training Loop}\label{alg:PKiL}
\begin{algorithmic}[1]
\While {loop over batch data samples $(u_i,y_i)$} \Comment $(u_i,y_i)$ is input pair indexed by $i$
\State Set input $u_i$ \Comment tokenized user document   
\State obtain $u_i^{pt}$ \Comment input after mapping to the Phrase Lexicon
\State obtain $\hat{u}_i^{pt} = EL({u}_i^{pt})$ \Comment EL represents embedding layer
\State obtain $S_{\hat{u_{i}}^{pt}} = SA(\hat{u}_i^{pt})$ \Comment Single Self-Attention layer 
\State $h_{\hat{u_{i}}^{pt}} = LN(S_{\hat{u_{i}}^{pt}})$ \Comment Layer normalization
\For{$k$, where $k \gets 1~to~K$} \Comment {$K$ is no. of process knowledge, and $EL_k$ is embedding layer for $k^{th}$ process knowledge block}
\State $Q = W_q^Th_{\hat{u_{i}}^{pt}}$, 
\State $K^{phq}_{k} = W_k^TEL_k(u_i^{p_k})$
\State $V^{phq}_{k} = W_v^TEL_k(u_i^{p_k})$
\State $z_{u_i} = softmax\bigg(\frac{QK^{phq}_{k}}{\sqrt{|h_{\hat{u_{i}}^{pt}}|}}V^{phq}_{k}\bigg)$ \Comment Cross Attention
\EndFor
\State $\hat{y_i}(u_i) = \sigma(\Theta^T(\oplus ~z_{u_i}))$ \Comment Row concat and Predict
\State If $L(\hat{y_i}(u_i),y_i(u_i)) \leq \varepsilon$, break \Comment {Convergence check}
\EndWhile
\end{algorithmic}
\end{algorithm}

\section{Datasets}
\noindent \textit{PRIMATE} \cite{gupta2022learning} and \textit{CLEF e-Risk 2021} (referred to as CLEF e-Risk) \cite{parapar2021overview} datasets were constructed with the primary purpose to seek the answer to following two specific questions regarding an LM: (a) Does an LM have an implicit capability to reach a correct outcome, without the need of explicit external knowledge? and (b) Does the attention component in LM focus on those phrases that can answer questions in CPG, used during annotation? Both datasets were created using the following format: (a) PRIMATE: $<title, post, \{Q_i:= yes/no\}_{i=1:9}>$ and (b) CLEF e-Risk: $<post,\{Q_i:= yes/no\}_{i=1:9}, D/\hat{D} >$; where D: Depression, and $\hat{D}$: Not-depression and $Q_i$ denotes a question in PHQ-9 questionnaire. PRIMATE comprises 2003 posts sourced from Reddit's r/depression\_help subreddit. In contrast, the CLEF e-Risk dataset is characterized by class imbalance and includes up to 2000 Reddit posts per user for a total of 828 users. Among these, 79 users have self-reported clinical depression, while 749 users serve as controls. The control group comprises random Redditors who are interested in discussing depression. Six MHPs affiliated with the National Institute of Mental Health and Neurosciences (NIMHANS) annotated both datasets in Bangalore, India. MHPs attained an inter-annotator agreement of 0.85 in PRIMATE and 0.77 in CLEF e-Risk using the Fleiss Kappa reliability measurement scale \cite{fleiss1971measuring}. 

Another resource considered in the current research is \textit{Causal Analysis of Mental health issueS (CAMS)} dataset \cite{garg2022cams}. CAMS differs from PRIMATE/CLEF e-Risk in two respects: (a) It focuses on causative triggers of depression; thus, the labels are causes of depression, and (b) the annotators explain each annotated sample by highlighting parts of the post. The dataset is provided in following format: $<post, \{C_i\}_{i=0:5}, \text{inference}>$; where $\{C_i\}$ represent six labels in following order: `No reason' ($C_0$), `Bias or abuse' ($C_1$), `Jobs and careers' ($C_2$), `Medication' ($C_3$), `Relationship' ($C_4$), and `Alienation' ($C_5$). Inference indicates the key phrases (or relevant concepts) in the post that expert annotators likely have considered when assigning labels. CAMS comprises 5051 posts with an inter-annotator agreement of 0.61 on the Fleiss Kappa scale. 
 
On the reviewer's comments, we expanded our datasets by adding the suicide risk dataset. 
\textbf{\textit{Reddit C-SSRS Dataset (R-CSSRS; Multi-Class):}} This is a unique dataset in comparison to PRIMATE and CLEF e-RISK. The ``C-SSRS" stands for Columbia Suicide Risk Severity Scale, a CPG to assess suicide risk \cite{posner2011columbia}. R-CSSRS was designed to assess the suicide risk severity of Reddit users who drift between different communities on Reddit to seek support. The dataset uses a suicide risk severity lexicon, which contains four categories: \{ suicide indicator, suicide ideation, suicide behavior, and suicide attempt.\} and an additional category: \{ supportive \}, which is for users who either have no signs of suicide risk or have recovered. R-CSSRS comprises 500 users with $\sim$7000 posts annotated by experts with an agreement score of 79\%. 

The datasets were divided into training, validation, and test sets using a 50:20:30 ratio. Stratified sampling was used for training and validation to maintain the target variable's distribution. Unlike standard cross-validation, stratified cross-validation ensures each fold represents the entire dataset, which is crucial for imbalanced classes. This approach allowed us to train and test the model multiple times with different data splits, improving performance and reliability and providing a more robust evaluation compared to a fixed split.

\noindent 
\textbf{\textit{Implementation and Training Details:}} \textit{CLEF e-Risk Dataset:} The \textbf{{\fontfamily{cmss}\selectfont PSAT}} model was trained with the following parameters: 2 classes (Depressed vs. Non-Depressed), a maximum text length of 1000, a training batch size of 32, a validation batch size of 32, 20 training epochs, a learning rate of 1e-03, and a random seed of 0. \textit{PRIMATE Dataset:} {\fontfamily{cmss}\selectfont PSAT} was trained with the following parameters: 9 classes (see Table 1), a maximum text length of 100, a training batch size of 32, a validation batch size of 32, 20 training epochs, a learning rate of 1e-03, and a random seed of 0. \textit{CAMS Dataset:} {\fontfamily{cmss}\selectfont PSAT} was trained with the following parameters: 6 classes (C0-C5), a maximum text length of 40, a training batch size of 256, a validation batch size of 256, 30 training epochs, a learning rate of 1e-03, and a random seed of 0. We used 'cross entropy' loss and the 'Adam' optimizer to train our models effectively for classification tasks. {\fontfamily{cmss}\selectfont PSAT} was trained using two NVIDIA Tesla V100 GPUs (16GB), and the training time was approximately 2 hours. 
{\fontfamily{cmss}\selectfont PSAT} code is publicly available for reproducibility at: \url{https://github.com/sumitnitkkr/PSAT}.

\section{Results and Analysis}
Results from {\fontfamily{cmss}\selectfont PSAT} on multiple depression-related datasets are compared with various domain-specific and general-purpose LMs, including domain-specific large language models such as MentalT5 \cite{Xu_2024}, MentalBART, MentaLLAMA 33B quantized using LORA \cite{yang2023mentalllama}.

\subsection{Evaluation Metrics} 
{\fontfamily{cmss}\selectfont PSAT} is quantitatively assessed using standard performance indicators, including Precision (P), Recall (R), Macro F1, and Matthew Correlation Coefficient (MCC) scores. 

Additionally, to assess the similarity between attention words generated by language models and the ground truth explanations in the CAMS dataset, we have assessed BERTScore \cite{zhang2019bertscore}. Finally, as a novel addition to evaluation methods, we introduce the Average Knowledge Capture (AKC) metric as a quantitative measure of explainability. 

\textit{\textbf{Average Knowledge Capture:}} AKC sums the cosine similarity between highlighted words/phrases and the concepts in PHQ-9 depression ontology, averaged over the number of posts and concepts in the PHQ-9 depression ontology. A higher AKC value suggests a strong semantic connection between attention in LMs and questions in CPGs, increasing the likelihood of addressing the questions within CPGs.  AKC is formulated as follows: 
\[
\frac{1}{|N_P||O|} \sum_{i \in N_P} \sum_{w \in P} \sum_{c \in O} \cos (z_{w_i}, z_c)\cdot \log(\cos(z_{w_i}, z_c))
\]
where $w$ and $c$ represent the number of words in a post $P$ and concepts in a class of the PHQ-9 depression ontology $O$, which represents a question in the PHQ-9 
respectively. $N_P$ is the total number of posts in the datasets. $z_{w_i}$ represents the embedding of the word/phrase marked as significant by attention-based LMs. $z_c$ represents the 300-dimension concept embedding. The AKC range is [0,1]. For readability, the scores in \autoref{tab:qresults} are multiplied by 100.

\begin{table}[t]
\begin{center}
\scriptsize
\resizebox{8cm}{!}{
\begin{tabular}{p{2cm}|ccccc}
    \toprule[1.5pt]
    \multirow{2}{*}{\Large{\textbf{Models}}} &\multicolumn{5}{c}{\textbf{CLEF e-Risk} (in \%)}\\ \cmidrule{2-6}
    & \textbf{P} & \textbf{R} & \textbf{F1 }& \textbf{MCC} & \textbf{AKC}($\uparrow$)  \\ \cmidrule{2-6}
    LongFormer & \underline{60.8} & \underline{48.7} & \underline{54.0} & \underline{36.1}  & \underline{7.7} \\
    RoBERTa & 31.4 & 25.1 & 27.8 & 17.3  & 3.0 \\
    BERT& 36.9 & 34.4 & 35.6  & 23.6 & 3.8 \\
    ERNIEv2 & 59.8 & 45.6 & 51.7 & 34.9 & 7.0 \\ \cline{2-6}
    MentalBERT & 53.5 & 50.6 & 52.0 & 35.5 & 7.2\\
    PsychBERT & 58.7 & 52.2 & 55.2 & 36.4 & 7.8 \\
    ClinicalT5 $\dagger$ & \underline{61.7} & \underline{54.8} & \underline{58.0} & \underline{39.9} & \underline{10.3} \\ 
    MentalT5 & 58.6 & 47.7 & 52.6 & 36.2 & 7.7 \\
    MentalBART & 60.9 & 55.8 & 58.2 & 39.7 & 9.8 \\  
    MentaLLAMA & 61.8 & 51.5 & 56.1 & 39.5 & 9.1 \\
    \cline{2-6}
    \textbf{{\fontfamily{cmss}\selectfont PSAT}}$\dagger$ & \textbf{63.4} & \textbf{55.7} & \textbf{59.3} & \textbf{44.4}  & \textbf{11.6} \\ \midrule[1pt]
    & \multicolumn{5}{c}{\textbf{PRIMATE} (in \%)} \\ \cmidrule{2-6}
    LongFormer & 49.9 & 41.3 & 45.1  & 17.7 & 13.8 \\
    RoBERTa & 54.2 & 51.5 & 52.8 & 31.4 & 16.7 \\
    BERT & \underline{58.7} & \underline{52.6} & \underline{55.4} & \underline{33.8} & \underline{17.7}\\
    ERNIEv2 & 54.8 & 52.0 & 53.3 & 30.2 & 16.5\\ \cline{2-6}
    MentalBERT & 56.8 & 47.3 & 51.6 & 16.2 & 15.0\\
    PsychBERT & 52.8 & 51.8 & 52.3 & 18.4 & 14.3\\
    ClinicalT5 $\dagger$ & \underline{59.4} & \underline{53.3} & \underline{56.2} & \underline{34.7} & \underline{19.6} \\ 
    MentalT5 & 54.6 & 51.8 & 53.2 & 33.8 & 18.7 \\
    MentalBART & 56.5 & 52.5 & 54.4 & 33.2 & 18.5 \\  
    MentaLLAMA & 60.1 & 53.2 & 56.4 & 34.2 & 19.2 \\
    \cline{2-6}
   {\fontfamily{cmss}\selectfont PSAT}$\dagger$ & \textbf{63.7} & \textbf{59.8} & \textbf{61.6} & \textbf{39.8} & \textbf{21.5} \\ \bottomrule[1.5pt]
\end{tabular}}
\end{center}
\caption{\footnotesize\textbf{Classification results on CLEF e-Risk and PRIMATE.} The best performance is indicated in bold, while the second best is underlined. The symbol $\uparrow$ next to AKC (Average Knowledge Capture) signifies that higher values are better in percentage units. $\dagger$ (p $<$ 0.05) indicates statistically significant results when comparing the best to the second-best metric.}
\label{tab:qresults}
\end{table}

\subsection{Classification and Explainability Performance}
\textit{\textbf{{\fontfamily{cmss}\selectfont PSAT} Performance on CLEF e-Risk Dataset:}} Table \ref{tab:qresults} reveals consistent improvement from {\fontfamily{cmss}\selectfont PSAT} over baseline LMs in the CLEF e-Risk dataset and other LMs like MentalT5, MentalBART and a large language model MentaLLAMA. Compared with the second-best \textit{general} LM, ClinicalT5 \cite{lu2022clinicalt5}, {\fontfamily{cmss}\selectfont PSAT} showed enhancements across various standard performance metrics. Specifically, {\fontfamily{cmss}\selectfont PSAT} exhibited 4.4\%, 6.6\%, 5.65\%, and 12\% improvements in precision, recall, F1, and MCC scores, respectively, compared to ClinicalT5, MentalBART, and MentaLLAMA. These models scored AKC values close to those of {\fontfamily{cmss}\selectfont PSAT}.

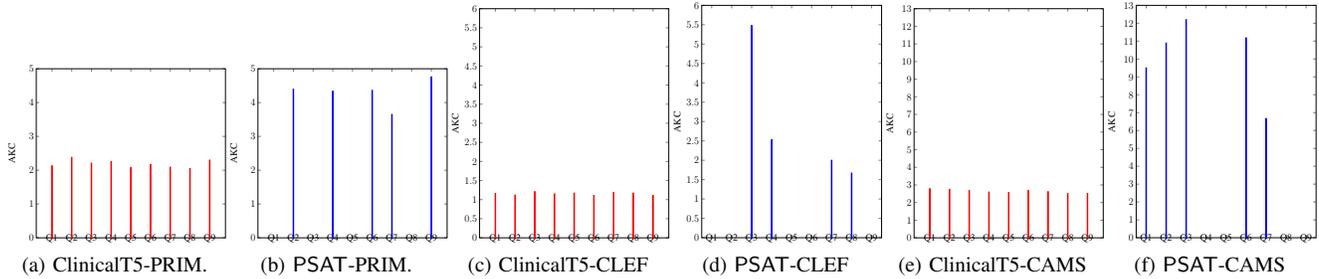
\begin{figure*}[!ht]

\begin{minipage}[h]{0.65\textwidth}
\subfloat[][ClinicalT5-PRIM.]{\resizebox{0.25\textwidth}{!}{
\begin{tikzpicture}
\tikzstyle{every node}=[font=\Large]
        \begin{axis}[x=1cm,
            ylabel={AKC},
            xticklabels={Q1,
                        Q2,
                        Q3,
                        Q4,
                        Q5,
                        Q6,
                        Q7,
                        Q8,
                        Q9},
            xtick=data, xticklabel style = {anchor = center},width=\textwidth,
                   bar width=0.7, ymin=0,
                    ymax=5.0,
          ]
            \addplot[ybar,fill=red!25, draw = red] coordinates {
                (1, 2.13)
                (2, 2.38)
                (3, 2.21)
                (4, 2.25)
                (5, 2.08)
                (6, 2.17)
                (7, 2.09)
                (8, 2.05)
                (9, 2.3)
            };
        \end{axis}
    \end{tikzpicture}}}
\subfloat[][\textbf{{\fontfamily{cmss}\selectfont PSAT}}-PRIM.]{\resizebox{0.25\textwidth}{!}{
\begin{tikzpicture}
\tikzstyle{every node}=[font=\Large]
        \begin{axis}[x=1cm,
            ylabel={AKC},
            xticklabels={Q1,Q2,Q3,Q4,Q5,Q6,Q7,Q8,Q9},
            xtick=data, xticklabel style = {anchor = center},width=\textwidth,
         bar width=0.7,
                    ymin=0,
                    ymax=5.0,
          ]
            \addplot[ybar,fill=blue!20, draw=blue] coordinates {

                (1, 0.0)
                (2, 4.393)
                (3, 0.0)
                (4, 4.337)
                (5, 0.0)
                (6, 4.361)
                (7, 3.648)
                (8, 0.0)
                (9, 4.761)
            };
        \end{axis}
    \end{tikzpicture}}}
\subfloat[][ClinicalT5-CLEF]{\resizebox{0.25\textwidth}{!}{
\begin{tikzpicture}
\tikzstyle{every node}=[font=\Large]
        \begin{axis}[x=1cm,
            ylabel={AKC},
            xticklabels={Q1,
                        Q2,
                        Q3,
                        Q4,
                        Q5,
                        Q6,
                        Q7,
                        Q8,
                        Q9},
            xtick=data, xticklabel style = {anchor = center},width=1.3\linewidth,
                   bar width=0.7, ymin=0,
                    ymax=6.0,
          ]
            \addplot[ybar,fill=red!25, draw = red] coordinates {             
                (1, 1.16)
                (2, 1.12)
                (3, 1.21)
                (4, 1.15)
                (5, 1.17)
                (6, 1.11)
                (7, 1.19)
                (8, 1.17)
                (9, 1.11)
            };
        \end{axis}
    \end{tikzpicture}}}
\subfloat[][\textbf{{\fontfamily{cmss}\selectfont PSAT}}-CLEF]{\resizebox{0.24\textwidth}{!}{
\begin{tikzpicture}
\tikzstyle{every node}=[font=\Large]
        \begin{axis}[x=1cm,
            ylabel={AKC},
            xticklabels={Q1,Q2,Q3,Q4,Q5,Q6,Q7,Q8,Q9},xmin=0.5,xmax=9.5,
            xtick=data, xticklabel style = {anchor = center},
         bar width=0.7, ymin=0, ymax=6.0, width=1.3\textwidth,
          ]
            \addplot[ybar,fill=blue!20, draw=blue] coordinates {
                (1, 0.0)
                (2, 0.0)
                (3, 5.48)
                (4, 2.53)
                (5, 0.0)
                (6, 0.0)
                (7, 2.00)
                (8, 1.67)
                (9, 0.0)
            };
        \end{axis}
    \end{tikzpicture}}}
\subfloat[][ClinicalT5-CAMS]{\resizebox{0.25\textwidth}{!}{
\begin{tikzpicture}
\tikzstyle{every node}=[font=\Large]
        \begin{axis}[x=1cm,
            ylabel={AKC},
            xticklabels={Q1,
                        Q2,
                        Q3,
                        Q4,
                        Q5,
                        Q6,
                        Q7,
                        Q8,
                        Q9},
            xtick=data, xticklabel style = {anchor = center},width=1.3\textwidth,
                   bar width=0.7, ymin=0,
                    ymax=13.0,
          ]
            \addplot[ybar,fill=red!25, draw = red] coordinates {             
                (1, 2.79)
                (2, 2.75)
                (3, 2.68)
                (4, 2.60)
                (5, 2.57)
                (6, 2.69)
                (7, 2.62)
                (8, 2.51)
                (9, 2.51)
            };
        \end{axis}
    \end{tikzpicture}}}
\subfloat[][\textbf{{\fontfamily{cmss}\selectfont PSAT}}-CAMS]{\resizebox{0.24\textwidth}{!}{
\begin{tikzpicture}
\tikzstyle{every node}=[font=\Large]
        \begin{axis}[x=1cm,
            ylabel={AKC},
            xticklabels={Q1,Q2,Q3,Q4,Q5,Q6,Q7,Q8,Q9},xmin=0.5,xmax=9.5,
            xtick=data, xticklabel style = {anchor = center},
         bar width=0.7, ymin=0, ymax=13.0, width=1.3\textwidth,
          ]
            \addplot[ybar,fill=blue!20, draw=blue] coordinates {
                (1, 9.50)
                (2, 10.89)
                (3, 12.20)
                (4, 0.0)
                (5, 0.0)
                (6, 11.18)
                (7, 6.67)
                (8, 0.0)
                (9, 0.0)
            };
        \end{axis}
    \end{tikzpicture}}}
\end{minipage}
\caption{\footnotesize The \textbf{AKC Scores} attained by \textbf{PSAT on PRIMATE (PRIM), CLEF e-Risk (CLEF), and CAMS} demonstrate the impact of cross-attention blocks. \textbf{PSAT}'s higher range of AKC scores indicates its ability to utilize cross-attention effectively to exclude less relevant PHQ-9 questions, thus influencing the quality of explanations and the results.
}
\label{fig:attn_weights_plots}
\end{figure*}

In comparison with domain-specific and second-best ClinicalT5 \cite{lu2022clinicalt5}, {\fontfamily{cmss}\selectfont PSAT} also showed 
benefit, particularly in MCC (10\%) compared to precision (3\%), recall (2\%), and F1 (2\%) scores. Moreover, we obtained larger gains in AKC scores compared to general LongFormer and ClinicalT5. Despite ClinicalT5 and {\fontfamily{cmss}\selectfont PSAT} having similar AKC scores, Figure \ref{fig:attn_weights_plots}(c,d) revealed that {\fontfamily{cmss}\selectfont PSAT}'s attention concentrated on specific questions in PHQ-9, whereas ClinicalT5's attention scattered across all nine PHQ-9 questions.

\textit{{\fontfamily{cmss}\selectfont PSAT} on PRIMATE:} The PRIMATE dataset utilizes PHQ-9 questions as ground-truth labels, directly compatible with AKC scoring. Thus, the AKC score was generally higher than CLEF e-Risk. Furthermore, 
most LMs 
emphasized words or phrases in the text pertinent to PHQ-9 questions. AKC scores for {\fontfamily{cmss}\selectfont PSAT} improved 23\% over BERT and 9\% over ClinicalT5 (see Table \ref{tab:qresults}). On standard metrics, {\fontfamily{cmss}\selectfont PSAT} provided 8\%, 12\%, 10\%, and 15\% gain over BERT and 7\%, 11\%, 9\%, and 13\% gain over ClinicalT5, in terms of precision, recall, F1, and MCC scores respectively. Further, {\fontfamily{cmss}\selectfont PSAT} reported average improvements of 11.8\%, 14\%, 12.75\%, and 18\% over MentalT5, MentalBART, and MentaLLAMA in terms of precision, recall, F1, and MCC scores, respectively. As in Figures \ref{fig:attn_weights_plots}(a,b) and \ref{fig:attn_weights_plots}(c,d) ClinicalT5 did direct 
marginally high attention on questions \{Q2, Q4, Q6, Q9\} in PRIMATE and \{Q4, Q7, Q8\} in CLEF e-Risk. However, {\fontfamily{cmss}\selectfont PSAT} magnified those scores 
by ruling out less relevant PHQ-9 questions.

\begin{table}[t]
\centering
\resizebox{8cm}{!}{
\begin{tabular}{p{2cm}cccccccc}
    \toprule[1.5pt]
    \multirow{2}{*}{\textbf{Models}} &\multicolumn{6}{c}{\textbf{F1-Score}} & \multirow{2}{*}{\textbf{BScore}} & \multirow{2}{*}{\textbf{AKC}}\\ \cmidrule{2-7}
    & $C_0$ & $C_1$ & $C_2$ & $C_3$ & $C_4$ & $C_5$ & & \\ \cmidrule{2-7}
    LR & 0.63 & 0.28 & 0.54 & 0.46 & 0.46 & 0.53 & 0.26 & 6.4 \\
    CNN-LSTM & 0.54 & 0.22 & 0.54 & 0.47 & 0.54 & 0.47 & 0.23 & 6.1 \\ \cline{2-9}
    ClinicalT5$\dagger$ & 0.71 & \underline{0.86} & 0.80 & \underline{0.73} & \underline{0.79} & \underline{0.82} & \underline{0.39} & \underline{23.7} \\ 
    MentalT5 & 0.66 & 0.79 & 0.72 & 0.73 & 0.68 & 0.77 & 0.33 & 22.0 \\ 
    MentalBART & 0.74 & 0.72 & \underline{0.84} & 0.57 & 0.64 & 0.67 & 0.33 & 20.8 \\ 
    MentaLLAMA$\dagger$ & \underline{0.78} & 0.82 & \underline{0.84} & 0.69 & 0.74 & \underline{0.82} & 0.38 & 23.0 \\ \cline{2-9}
    {\fontfamily{cmss}\selectfont {\fontfamily{cmss}\selectfont PSAT}}$\dagger$ & 0.83 & 1.0 & 0.83 & 0.66 & 0.83 & 0.83 & 0.47 & 50.4\\ \bottomrule[1.5pt]
\end{tabular}}
\caption{\textbf{Results on CAMS dataset}. $\dagger$ (p $<$ 0.05) indicates the results are statistically significant when comparing the best to the second-best metric. BScore: BERTScore, AKC: Average Knowledge Capture, LR: Logistic Regression, CNN-LSTM: spatial Convolutional Neural Network (CNN) with sequential Long Short Term Network (LSTM). It is worth noting that LR and CNN-LSTM were reported as strong baselines in CAMS \cite{garg2022cams}.}
\label{tab:qresults_cams}
\end{table}

\begin{table}[!ht]
    \scriptsize
    \centering
    \begin{tabular}{c|c|c|c}
    \toprule[1.5pt]
         Models & Accuracy & AUC-ROC & AKC ($\uparrow$) \\ \midrule[1pt]
         RoBERTa \cite{roy2022proknow} & 70.7 & 62.3 & 26.6 \\
         Longformer(2048) \cite{roy2022proknow} & 67.5 & 48.4 & 19.8 \\ \cline{2-4}
         ClinicalT5 & 71.0 & 62.5 & 25.0\\
         MentalT5 & 72.5 & 62.7 & 29.0 \\
         MentalBART & 72.4 & 62.2 & 27.5 \\
         MentaLLAMA & 73.3 & 63.0 & 20.0\\ \cline{2-4}
        {\fontfamily{cmss}\selectfont PSAT} & \textbf{72.1} &\textbf{63.2} & \textbf{32.4} \\ \bottomrule[1.5pt] 
    \end{tabular}
    \caption{Scores reported on CSSRS. Results are in percentages. }
    \label{tab:cssrs}
\end{table}

\textit{{\fontfamily{cmss}\selectfont PSAT} on CSSRS:} To evaluate the generalizable effectiveness of {\fontfamily{cmss}\selectfont PSAT} in assessing suicide severity within the R-CSSRS dataset, we reduced the cross-attention blocks from 9 to 5, focusing on five categories: supportive, suicide indicator, suicide ideation, suicide behavior, and suicide attempt. The {\fontfamily{cmss}\selectfont PSAT}  results, presented in \autoref{tab:cssrs}, are compared only with RoBERTa and Longformer, as these models yielded acceptable performance on the R-CSSRS dataset \cite{roy2022process}. Our experiments demonstrate that integrating CSRRS CPG not only improves performance but also enables the model to base outcomes on clinical concepts rather than speculative parametric knowledge, as shown by MentaLLAMA with an AKC score of 20.0.

On the CSSRS dataset, {\fontfamily{cmss}\selectfont PSAT} demonstrated significant improvements across various metrics compared to other models. Specifically, {\fontfamily{cmss}\selectfont PSAT} showed a 1.98\% improvement in accuracy, 1.44\% in AUC-ROC, and 21.80\% in AKC over RoBERTa. Compared to Longformer, {\fontfamily{cmss}\selectfont PSAT} achieved a 6.81\% increase in accuracy, 30.58\% in AUC-ROC, and 63.64\% in AKC. When compared to ClinicalT5, {\fontfamily{cmss}\selectfont PSAT} improved accuracy by 1.55\%, AUC-ROC by 1.12\%, and AKC by 29.60\%. In comparison with MentalT5, {\fontfamily{cmss}\selectfont PSAT} showed a 0.80\% increase in AUC-ROC and an 11.72\% increase in AKC, despite a slight reduction in accuracy by 0.55\%. Similarly, against MentalBART, {\fontfamily{cmss}\selectfont PSAT} achieved a 1.61\% improvement in AUC-ROC and a 17.82\% increase in AKC, with a minor decrease in accuracy by 0.41\%. Finally, when compared to MentaLLAMA, {\fontfamily{cmss}\selectfont PSAT} demonstrated a 0.32\% increase in AUC-ROC and a 38.00\% improvement in AKC, although it had a small reduction in accuracy by 1.64\%. The paradoxical decline in AKC, despite improvements in AUC-ROC and Accuracy for MentaLLAMA, can be attributed to the excessive verbosity in the generative explanations. The excessive verbosity negatively impacts AKC, which measures the number of relevant concepts per sentence in explanations. This shows that despite extensive fine-tuning on mental healthcare datasets, MentaLLAMA and MentalT5 underperformed compared to {\fontfamily{cmss}\selectfont PSAT}. This is likely because these domain-specific models weren't trained on suicide-specific data. In contrast, {\fontfamily{cmss}\selectfont PSAT}, even without fine-tuning on domain-specific datasets, proved effective for sensitive mental health disorders like suicide.



\textit{\textbf{{\fontfamily{cmss}\selectfont PSAT} on CAMS:}} CAMS provided ``inference,'' reported by annotators as an explanation behind their labeling. Such ground-truth explanations enable a more thorough assessment of the fidelity of the {\fontfamily{cmss}\selectfont PSAT's} attention allocation. 
Table \ref{tab:qresults_cams} shows the performance gains of {\fontfamily{cmss}\selectfont PSAT} over CAMS baselines and ClinicalT5. As Logistic Regression (LR), Convolutional Neural Network (CNN), and Long Short Term Memory (LSTM) lack the ``attention" component found in {\fontfamily{cmss}\selectfont PSAT} or ClinicalT5, we computed BERTScores for measuring the similarity between the high-ranked features and ground-truth explanations in CAMS. For feature importance, we used LIME and created a rank list of features, setting a cut-off of 0.4. We kept such a low threshold to enable the inclusion of numerous features that can contribute to the model's certainty in the classification. {\fontfamily{cmss}\selectfont PSAT} demonstrated notable improvements of 45\%, 51\%, and 17\% over LR, CNN+LSTM, and ClinicalT5, respectively, using BERTscore. Using AKC\footnote{AKC scores for LR and CNN+LSTM were computed by creating embeddings of features using BERT}, {\fontfamily{cmss}\selectfont PSAT's} attention remains focused on a 
few PHQ-9 questions compared to LR, CNN+LSTM, and ClinicalT5 (see Figure \ref{fig:attn_weights_plots}(e,f)).

\subsection{Explainability Results}
\begin{figure}[t]
\centering
\includegraphics[width=0.5\textwidth]{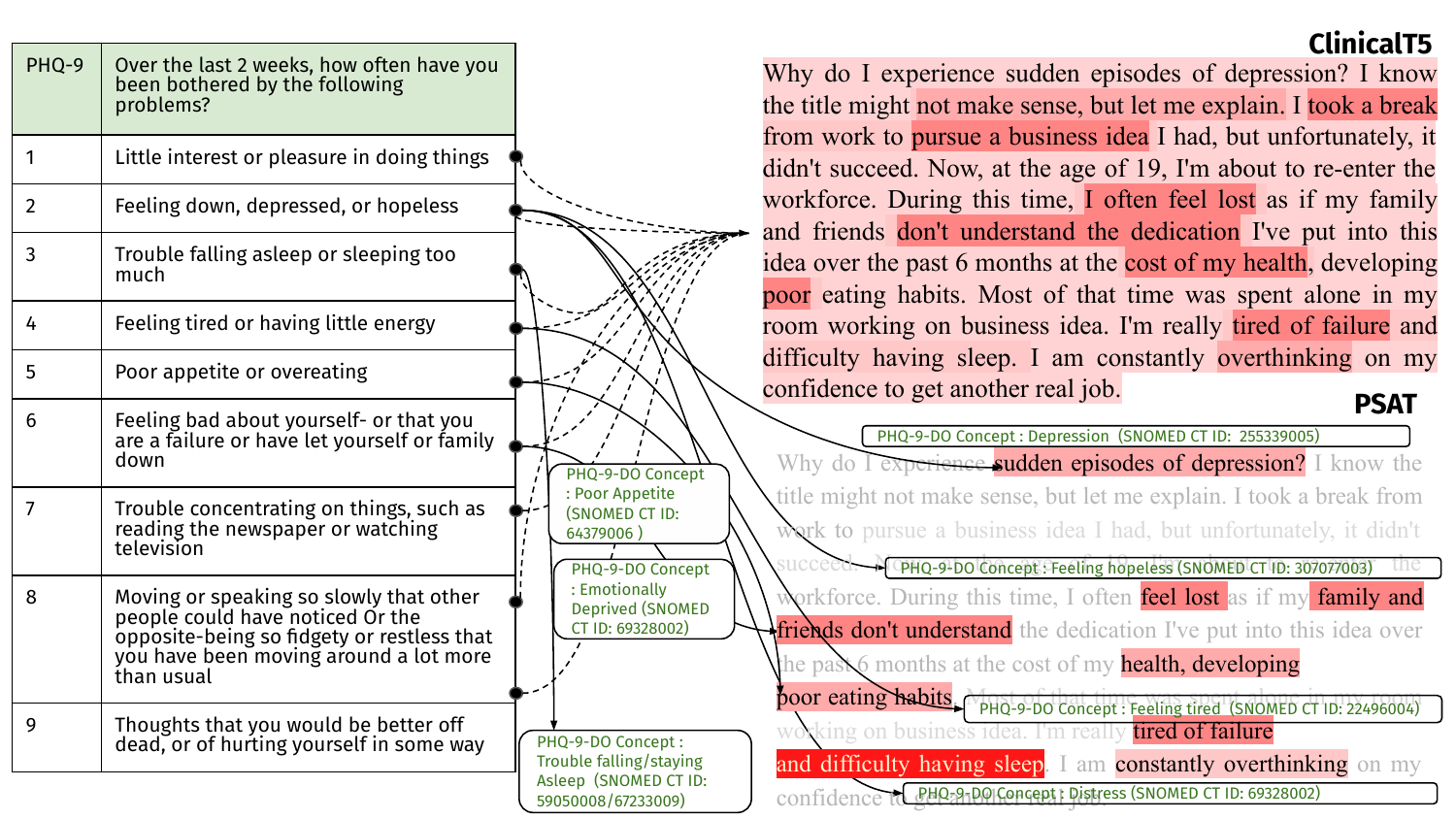}
\caption{\footnotesize Illustration of the process followed by PHQ-9 Class Visualizer in mapping attention words of {\fontfamily{cmss}\selectfont PSAT} to PHQ-9-DO. The SNOMED-CT IDs in round boxes demonstrate a trace to clinically grounded concepts. These mapped concepts are then utilized to produce explanations as presented in Table \ref{tab:hlex}. Additionally, a contrast is drawn by showcasing ClinicalT5's attention, which highlights the entire text and gives low weight to important clinical concepts. More examples of explanations are provided in \textit{\textbf{supplementary material}} from CLEF eRisk (page 2), CAMS (pages: 3-7), and PRIMATE (pages: 8-11). }
\label{fig:explanations}
\end{figure}

We prompted GPT 3.5 to generate explanations. The method inputs the post and word/phrases from ClinicalT5 or concepts from {\fontfamily{cmss}\selectfont PSAT}. We use the ``text-DaVinci-003" model of GPT 3.5 in the Python library called LangChain, created by 
\cite{Chase2023langchain}. The concepts provided by {\fontfamily{cmss}\selectfont PSAT} differed from those of ClinicalT5, while GPT 3.5 uniquely incorporated them during explanation generation (see italicized words in Table \ref{tab:hlex}). However, words from the ClinicalT5 model did not result in a similar pattern in GPT 3.5's behavior. Figure \ref{fig:explanations}, shows that ClinicalT5 highlights the entire text with nearly equal probability, 
failing to distinguish which specific words GPT 3.5 focused on during its explanation generation process. 


\begin{table}[!t]
    \centering
    \scriptsize
    \begin{tabular}{p{8.5cm}}
    \begin{tcolorbox}[colback=white,
                  colframe=black,
                  width=8.5cm,
                  arc=1mm auto outer arc, title={Input Text},colbacktitle=white, coltitle=black
                 ]
         Why do I experience sudden episodes of depression? I know the title might not make sense, but let me explain. I took a break from work to pursue a business idea I had, but unfortunately, it didn't succeed. [...] During this time, I often feel lost as if my family and friends don't understand the dedication I've put into this idea over the past 6 months at the cost of my health, developing poor eating habits. [...] I'm really tired of failure and difficulty having sleep. I am constantly overthinking.
        \end{tcolorbox}\\ 
         \begin{tcolorbox}[colback=white,
                  colframe=black,
                  width=8.5cm,
                  arc=1mm, auto outer arc, title={ClinicalT5 on Words/Phrases},colbacktitle=white, coltitle=black
                 ]
        The text describes sudden episodes of depression caused by investing so much time and effort into pursuing a business idea that didn't work out as planned. The speaker's feelings are valid, and it's okay to feel this way. It's not uncommon for young entrepreneurs to face setbacks and challenges, and when their endeavors don't succeed, it can be emotionally taxing. The speaker feels lost with a lack of support from family and friends.
        \end{tcolorbox}\\ 
         \begin{tcolorbox}[colback=white,
                  colframe=black,
                  width=8.5cm,
                  arc=1mm, auto outer arc, title={{\fontfamily{cmss}\selectfont PSAT} on Concepts},colbacktitle=white, coltitle=black
                 ]
        The text maps to a person who is \textit{depressed}, \textit{feeling hopeless}, and lost. They took a break from work to pursue a failed business idea, leading to feelings of disappointment and \textit{feeling of tiredness}. During the pursuit of this idea, they sacrificed their health, resulting in \textit{poor appetite} and difficulties with \textit{falling asleep} and \textit{staying asleep}. The person also feels isolated as friends and family are \textit{emotionally unavailable}. The constant overthinking about their confidence to find another job adds to their \textit{distress}.
        \end{tcolorbox}
    \end{tabular}
 \caption{\footnotesize \textbf{Illustrative instances showcasing the quality of explanations from {\fontfamily{cmss}\selectfont PSAT} compared to ClinicalT5, employing GPT3.5.} {The explanations generated by {\fontfamily{cmss}\selectfont PSAT} exhibit an application-relevant explainability as they accurately capture the context within the user's expression by using concepts in PHQ-9-DO. Conversely, other GPT3.5(ClinicalT5) explanations appear to deviate from the intended context.}}
    \label{tab:hlex}
\end{table}

\noindent \textit{\textbf{Application-Relevant Explainability using PHQ-9 Class Visualizer:}}  
We establish a connection between attention words and concepts in PHQ-9-DO by computing their cosine similarity. The visualizer then extracts top-ranked concept(s) with a similarity greater than 0.80, considering them potential concepts for generating explanations using GPT 3.5. GPT 3.5 utilized the mapped concepts, as illustrated in Figure \ref{fig:explanations}, to generate an explanation, as shown in Table \ref{tab:hlex}. The presence of SNOMED-CT IDs with each concept is a trace to an MHP-understandable knowledge source, reinforcing confidence in {\fontfamily{cmss}\selectfont PSAT's} outcomes.

\noindent \textit{\textbf{Comparison with Human Explanation:}} Five faculty members from the Department of Psychology at the University of Maryland provided explanations for 15 posts annotated with clinical labels.  We then compared their explanations with {\fontfamily{cmss}\selectfont PSAT}  explanations using Linguistic Inquiry Word Count (LIWC) software \cite{boyd2022}. Across all sample pairs, no statistically significant \textit{general} lexical or syntactic content favored {\fontfamily{cmss}\selectfont PSAT} explanations as measured by word count, clout, authenticity, tone, words per sentence, or linguistic properties including function words and pronouns. However, and most importantly, the LIWC metric ``analytic" indicates a statistically significant advantage for {\fontfamily{cmss}\selectfont PSAT}  t(14)= 5.47 \( p < .001\).

AKC metrics on clinician-generated and {\fontfamily{cmss}\selectfont PSAT}-generated explanations are relatively high  ($\bar{x}$ clinician = .346; $\bar{x}$ {\fontfamily{cmss}\selectfont PSAT} = .350). The average mean difference is .0042 (or .42 \%), which favors {\fontfamily{cmss}\selectfont PSAT}.  This is a trivial difference relative to the magnitude of differences between computational models over many more examples. Nevertheless, a dependent t does suggest an advantage for {\fontfamily{cmss}\selectfont PSAT} over clinical explanations because the variance in these selected items is so small t(14)= 3.233 \( p < .006\).

We conclude that the content of {\fontfamily{cmss}\selectfont PSAT} explanations is at least as good as human explanations.

\section{Discussion \& Conclusion}
The limitations of ``black box'' LMs have hindered their widespread use in healthcare. We explored whether it is feasible to design an architecture like {\fontfamily{cmss}\selectfont PSAT}  that is inherently explainable, particularly through the lens of CPGs and related clinical terms. In response, we introduced the {\fontfamily{cmss}\selectfont PSAT}  model, an innovative cross-attention LM that integrates process knowledge—like a depression severity assessment tool (PHQ-9)—to provide relevant explanations while detecting depression disorder. This approach not only makes explanations an integral part of the process but also adapts well to various applications as the architecture is designed to incorporate KBs. Our {\fontfamily{cmss}\selectfont PSAT} architecture enhanced performance across three datasets by improving how it handles complex cases, such as the CAMS dataset that includes explicit explanations. Designing an architecture that integrates clinical knowledge (such as CPGs) into the LM makes it better equipped to discern varied symptoms and experiences in unstructured text. This expansion of learning capabilities allows LMs to more effectively support mental health diagnosis, treatment, and intervention strategies. We also showed how {\fontfamily{cmss}\selectfont PSAT}  excels in extracting and tagging key phrases from diverse sources of unstructured, labeled mental health texts. By replacing standard transformer components with cross-attention mechanisms, our model infuses knowledge, enabling logical reasoning and continuous learning, which are crucial for developing robust and error-resistant LMs (as shown in Tables \ref{tab:qresults} and \ref{tab:qresults_cams}). Specifically, in Table \ref{tab:qresults_cams}, our {\fontfamily{cmss}\selectfont PSAT}  architecture has significantly progressed in handling the challenging CAMS dataset containing ground-truth explanations. Moreover, we facilitate collaborative decision-making by fostering a shared semantic understanding between LMs and human experts. The challenge in medical diagnosis involves sorting patient data into categories of co-morbidities, a task complicated by the overlapping nature of symptoms and conditions. Acknowledging and modeling these relationships can lead to more informed decision-making, heightened efficiency, and better outcomes in healthcare settings where domain-specific knowledge is critical.

\textit{\textbf{Findings:}}
\begin{itemize}
\item The generalizable architecture of {\fontfamily{cmss}\selectfont PSAT} leads to superior performance in detecting depression using PRIMATE and CAMS datasets, outperforming fine-tuned and general-purpose models. {\fontfamily{cmss}\selectfont PSAT's}  effectiveness is attributed to its ability to focus on clinically relevant concepts.
\item Incorporating clinical practice guidelines as questions for attention checking significantly improves the identification of key concepts over less important words, grounding the decision-making process of models. 
\item As demonstrated through attention analysis, fine-tuned language models lack explainability compared to {\fontfamily{cmss}\selectfont PSAT}. {\fontfamily{cmss}\selectfont PSAT}  uses BERT, which is not only a simple language model but also transparent compared to other domain-specific language models and MentaLLAMA. We found that knowledge infusion at the attention layer helps the model steer its attention to meaningful concepts.  
\item The addition of an experiment on suicide risk severity detection using the CSSRS questionnaire, IJBHI reviewers suggested that this further demonstrates the generalizability of the proposed approach.
\end{itemize}
 \noindent \textbf{\textit{Impact:}}  
This research expands the potential of knowledge infusion, where state-of-the-art LMs are complemented with external domain knowledge to provide clinically relevant explainability. 
{\fontfamily{cmss}\selectfont PSAT} takes us closer to classifying at-risk users in online communities using established clinical guidelines. The combined influence of social media on MH, the popularity of MH forums, and the increasing prevalence of MH concerns, particularly in adolescents and young adults, reinforces the necessity for valid screening tools. 

\textbf{\textit{Limitations:}}
We focused exclusively on text data from Reddit, which may limit the scope of our findings. Additionally, the CAMS dataset restricts us to six classes of depression, which may not encompass the full spectrum of the condition. Different conditions may necessitate distinct sets of guidelines and knowledge bases for knowledge infusion. It is essential to thoroughly investigate which sources are suitable for incorporation into the model to ensure meaningful and accurate results. Any gaps or inaccuracies in these knowledge bases can directly impact the model's performance and the relevance of the generated explanations. 
Comparison of human-generated and PSAT-generated explanations for user acceptance merits additional detailed, systematic investigation. Substantive differences in specific MH conditions and output style may contribute to user acceptance in ways that necessitate additions to PSAT capabilities. This is the focus of ongoing research.
Finally, the model is designed to assist MHPs rather than replace them. There is a risk that over-reliance on AI tools could lead to reduced human oversight, which is particularly critical in mental health diagnoses and interventions.

\noindent \textbf{\textit{Ethical Statement:}} Emphasizing the sensitive nature of this work, {\fontfamily{cmss}\selectfont PSAT} was trained and tested on datasets obtained through proper channels governed by dataset authors. {\fontfamily{cmss}\selectfont PSAT} does not offer a depression/suicide diagnostic conclusion. Our models and analyses should be one component of a distributed MHP-in-the-loop system for risk assessment. Given the criticality and subjectivity of an MH diagnosis, we performed two rounds of evaluations from MHPs. As {\fontfamily{cmss}\selectfont PSAT} is a tool to assist MHPs, it requires an extension for more complex differential diagnostic processes (e.g., for conditions such as Post-Traumatic Stress Disorder and postpartum depression). \textbf{Bias Mitigation:} We have implemented various strategies to mitigate bias in our AI models, including using diverse datasets for training and employing techniques to ensure fair and unbiased predictions across different demographic groups. We continually evaluate our models to identify and rectify any emerging biases. While diverse datasets help prevent bias, knowledge bases (KBs) play a crucial role in this mitigation. However, addressing bias within the KBs themselves remains an important challenge.
\textbf{Privacy:} Protecting patient privacy is paramount. Our models are designed to handle data in a way that ensures anonymity and confidentiality. We adhere to strict data protection protocols and comply with relevant regulations such as GDPR and HIPAA. Additionally, we only use de-identified data for training and validation purposes.
\textbf{Potential Harm to Vulnerable Populations:} We recognize the potential risks AI models may pose to vulnerable populations. To mitigate these risks, we provide explanations to ensure that our models are used as decision-support tools rather than standalone diagnostic instruments. This approach emphasizes the importance of human oversight by mental health professionals. We also conduct regular assessments to identify and address any unintended negative impacts on vulnerable groups.



\appendices


\section*{Acknowledgment}
We acknowledge partial support from the NSF EAGER award \#2335967, the UMBC SURFF Award, and the University Grant Commission (UGC) of India. Any opinions, conclusions, or recommendations expressed in this material are those of the authors and do not necessarily reflect the views of the NSF, UMBC, or UGC.

%
%

\bibliographystyle{IEEEtran}
\bibliography{manuscript}

\end{document}